%% file: MIMO.tex
\documentclass[letterpaper]{article} 
\usepackage{aaai23}  
\usepackage{times}  
\usepackage{helvet}  
\usepackage{courier}  
\usepackage[hyphens]{url}  
\usepackage{graphicx} 
\usepackage{natbib}  
\usepackage{caption} 
\usepackage{algorithm}
\usepackage{algorithmic}

\usepackage{newfloat}
\usepackage{listings}
\DeclareCaptionStyle{ruled}{labelfont=normalfont,labelsep=colon,strut=off} 
\floatstyle{ruled}
\newfloat{listing}{tb}{lst}{}
\floatname{listing}{Listing}
\usepackage{graphicx}
\usepackage{amsmath}
\usepackage{amssymb}
\usepackage{booktabs}
\usepackage{comment}
\usepackage{subcaption}
\usepackage[capitalize]{cleveref}
\crefname{section}{Sec.}{Secs.}
\Crefname{section}{Section}{Sections}
\Crefname{table}{Table}{Tables}
\crefname{table}{Tab.}{Tabs.}

\usepackage{enumitem}
\setlist[itemize]{noitemsep,nolistsep}
\usepackage{multirow}
\usepackage{arydshln}
\newcommand{\ie}{\textit{i}.\textit{e}.}

\title{MIMO Is All You Need : A Strong Multi-in-Multi-Out Baseline \\for Video Prediction}
\author{
    Shuliang Ning\textsuperscript{\rm 1,\rm 2}\equalcontrib, 
    Mengcheng Lan\textsuperscript{\rm 2}\equalcontrib, 
    Yanran Li\textsuperscript{\rm 3},  
    Chaofeng Chen\textsuperscript{\rm 4},
    Qian Chen\textsuperscript{\rm 5}, \\
    Xunlai Chen\textsuperscript{\rm 5\dag}, 
    Xiaoguang Han\textsuperscript{\rm2,\rm 1}\thanks{Both are Corresponding Authors.},
    Shuguang Cui\textsuperscript{\rm 2,\rm 1}}
\affiliations{
    \textsuperscript{\rm 1} FNii, CUHKSZ  \\
    \textsuperscript{\rm 2} SSE, CUHKSZ \\ 
    \textsuperscript{\rm 3} The University of Edinburgh\\
    \textsuperscript{\rm 4} Nanyang Technological University \\
    
    \textsuperscript{\rm 5} Shenzhen Meteorological Bureau\\

    shuliangning@link.cuhk.edu.cn, \{lanmengchengds, chaofenghust\}@gmail.com, yli19@ed.ac.uk, \\  \{chenqian, chenxunlai\}@weather.sz.gov.cn, \{hanxiaoguang, shuguangcui\}@cuhk.edu.cn 
}

\usepackage{graphicx}
\usepackage{amsmath}
\usepackage{amssymb}
\usepackage{booktabs}
\usepackage{comment}
\usepackage{subcaption}
\usepackage[capitalize]{cleveref}
\crefname{section}{Sec.}{Secs.}
\Crefname{section}{Section}{Sections}
\Crefname{table}{Table}{Tables}
\crefname{table}{Tab.}{Tabs.}

\usepackage{enumitem}
\setlist[itemize]{noitemsep,nolistsep}
\usepackage{multirow}
\usepackage{arydshln}

\usepackage{algorithm}
\usepackage{algorithmic}

\usepackage{newfloat}
\usepackage{listings}
\begin{document}
\maketitle
\begin{abstract}
\input{PaperBody/abstract}
\end{abstract}

\section{Introduction}
\label{sec:introduction}

\input{PaperBody/introduction}

\section{Related Work}
\label{sec:relatedwork}
\input{PaperBody/relatedwork4}

\section{Proposed Method}
\label{sec:method}
\input{PaperBody/method2}

\section{Experimental Results}
\label{sec:experiment}
\input{PaperBody/experiment3}

\section{Conclusion}
\label{sec:conclusion}
\input{PaperBody/conclusion3}

\section*{Acknowledgments}
This work is supported by National Key Research and Development Program of China for Intergovernmental Cooperation with grant No. 2019YFE0110100, the Basic Research Project No. HZQB-KCZYZ-2021067 of Hetao Shenzhen-HK S\&T Cooperation Zone, NSFC with Grant No. 62293482, the National Key R\&D Program of China with grant No. 2018YFB1800800, the Shenzhen Outstanding Talents Training Fund 202002, the Guangdong Research Projects No. 2017ZT07X152 and No. 2019CX01X104, the Guangdong Provincial Key Laboratory of Future Networks of Intelligence (Grant No. 2022B1212010001), the Shenzhen Key Laboratory of Big Data and Artificial Intelligence (Grant No. ZDSYS201707251409055), Science and technology innovation team project of Guangdong Meteorological Bureau No.GRMCTD202104, and Innovation and Development Project of China Meteorological Administration (Grant No. CXFZ2022J002). Thanks Oisin Mac Aodha for his advices on our work.

\bibliography{aaai23_MIMO}

\end{document}

%% file: PaperBody/abstract.tex
The mainstream of the existing approaches for video prediction builds up their models based on a Single-In-Single-Out (SISO) architecture, which takes the current frame as input to predict the next frame in a recursive manner. This way often leads to severe performance degradation when they try to extrapolate a longer period of future, thus limiting the practical use of the prediction model. Alternatively, a Multi-In-Multi-Out (MIMO) architecture that outputs all the future frames at one shot naturally breaks the recursive manner and therefore prevents error accumulation. However, only a few MIMO models for video prediction are proposed and they only achieve inferior performance due to the date. 
The real strength of the MIMO model in this area is not well noticed and is largely under-explored. Motivated by that, we conduct a comprehensive investigation in this paper to thoroughly exploit how far a simple MIMO architecture can go. Surprisingly, our empirical studies reveal that a simple MIMO model can outperform the state-of-the-art work with a large margin much more than expected, especially in dealing with long-term error accumulation. 
After exploring a number of ways and designs, we propose a new MIMO architecture based on extending the pure Transformer with local spatio-temporal blocks and a new multi-output decoder, namely MIMO-VP, to establish a new standard in video prediction. We evaluate our model in four highly competitive benchmarks. 
Extensive experiments show that our model wins 1st place on all the benchmarks with remarkable performance gains and surpasses the best SISO model in all aspects including efficiency, quantity, and quality. A dramatic error reduction is achieved when predicting 10 frames on Moving MNIST and Weather datasets respectively. We believe our model can serve as a new baseline to facilitate the future research of video prediction tasks. The code will be released.

%% file: PaperBody/introduction.tex
\begin{figure}[!t]
	\centering
	\includegraphics[width=1\linewidth]{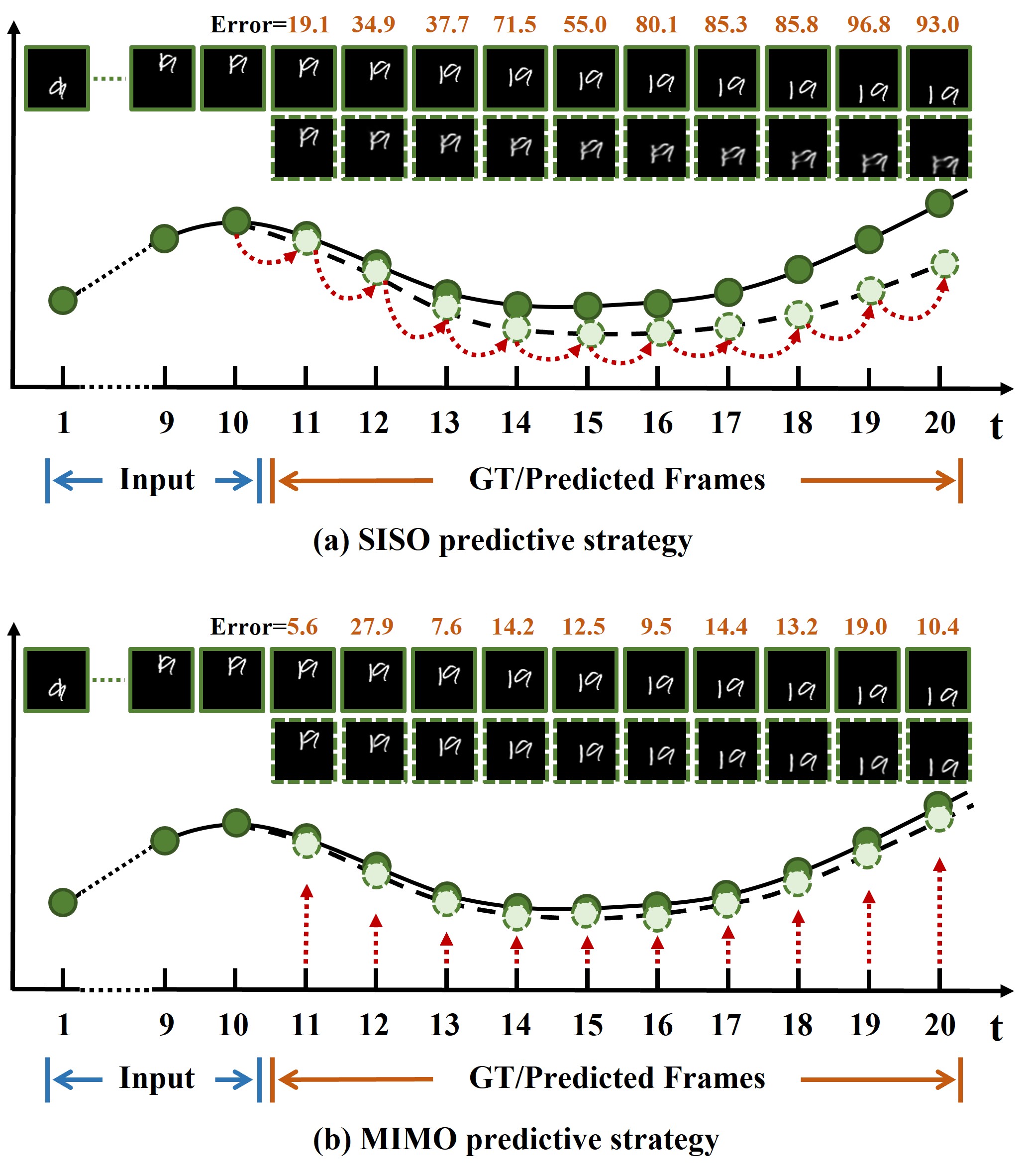}
	\caption{Illustration of MIMO and SISO on Moving MNIST: The Errors are from (a) PhyDNet \cite{guen2020disentangling} (b) Our model.	Figures with solid and dashed frames denote the ground-truth and predicted images respectively. The red arrows indicate the different error propagation way.
	}
	\label{fig:onecol}
\end{figure}

Video prediction aims to forecast future frames of a video sequence conditioned on previous frames. Since anticipating the future is such a fundamental capability for intelligent agents, video prediction has received increasing attention and benefited many applications such as precipitation forecasting~\cite{xingjian2015convolutional}, autonomous driving~\cite{kwon2019predicting}, and action recognition~\cite{liu2017video}. In the practical scenario of these real-life applications, an accurate prediction of the longer-term future will be greatly beneficial.

However, the current approaches are quite limited on the prediction horizon. Most of them develop sophisticated models based on Single-In-Single-Out (SISO) architecture such as Recurrent Neural Networks~\cite{wang2017predrnn,wang2018predrnn++,wang2019memory,guen2020disentangling,Yu2020EfficientAI,lin2020self,wu2021motionrnn,lee2021video}, Convolutional Long Short-Term Memory (ConvLSTM)~\cite{xingjian2015convolutional} and their variants. These recurrent models learn a hidden state for the history information and generate the next frame conditioned on the current predicted frame. 
As a result, their quality and accuracy will degrade quickly when the generated frames go longer due to the Butterfly Effects of the small error produced in the earlier frames. Especially, video data is high-dimensional and their models have great complexity, the small errors are easy to be amplified into serious compound errors over time. Some hierarchical models predict the low-dimensional structures of the frames firstly and transfer them back into pixel space to reduce the error accumulation. 
They either extract the human pose keypoints~\cite{kim2019unsupervised, walker2017pose, villegas2017learning}, face landmarks~\cite{yang2018pose, yan2018mt} or semantic label maps~\cite{lee2021revisiting}. With the constraints of the high level structures, their methods can generate semantic reasonable motions in a very long time period. However, these methods are infeasible to be applied to unstructured video data such as weather and radar videos. More importantly, all of these methods only alleviate the problem by reducing the error generated at once but did not solve it essentially by breaking the error propagation chain.

In contrast, we notice that some approaches for prediction tasks build up their models in a Multi-In-Multi-Out (MIMO) manner which takes all the inputs into the encoder and output all the frames from the decoder at once, such as DVF \cite{liu2017video} and SimVP \cite{gao2022simvp}. Compared with SISO, their future predictions do not rely on the previous ones and errors are not accumulated. Nevertheless, only few attempts for video prediction take MIMO architecture and their performances remain inferior to the best SISO models \cite{chang2021mau}. \textit{ Have the ability of MIMO architecture been underestimated for video prediction?}


We are curious to figure out how far the MIMO architecture can achieve in video prediction. We implement MIMO architecture by a pure Transformer~\cite{vaswani2017attention} model which demonstrates great success in many vision tasks recently as a general backbone since it is superior to capturing the global dependencies of input frames and output frames. However, it is worth noting that simply applying the trending Transformer models or their variants (Swin Transformer~\cite{liu2021swin}, etc) is not able to deal with this problem successfully. Although Vision Transformer-based models~\cite{arnab2021vivit, bertasius2021space, liu2022video} have achieved promising performance in video recognition, their extensions~\cite{rakhimov2020latent, weissenborn2019scaling} for video prediction are very limited and only output a single frame at once. Our experiments show Vanilla Transformer model only leads to inferior model performance since they struggle to exploit two very important clues for videos -- \textit{the spatial dependencies of each frame and the complex spatio-temporal dynamics.} Notably, it is non-trivial to extend a basic Transformer from time series prediction problem for a spatio-temporal series prediction task. 

To fill this gap, our objective of this paper is to develop an effective MIMO architecture for video prediction by carrying out a series of operations to tailor the basic architecture. Due to our best knowledge, this perspective has never been investigated in-depth. Various MIMO architectures for video prediction are empirically compared and studied in this work. 

For the global spatio-temporal dependency clues,
we incorporate the 2D convolutional structure into the multi-head attention to simultaneously capture the long term temporal dependencies of sequence and preserve the spatial information of frames.
For the most important spatio-temporal clues, we design a Spatio-temporal block which replaces the simple forward layers both in the encoder and decoder. This block will capture the local Spatio-temporal context to benefit the prediction. While many concrete implementations of this block are possible to capture the local context, we reveal that a simple design of 3D convolutions layers is already capable to yield considerably performance gains than individual art methods. On Moving MNIST dataset, this block leads to a decrease of 16\% MSE error. 

Moreover, another problem is the decoder of the typical Transformer cannot achieve MIMO straightforwardly. To deal with that, we design a new Multi-Output decoder which takes all the placeholder of the output frames as input features. Then we rectify the last layer of the decoder to generate the output frames in one shot. This decoder captures more dependencies of the subsequent frames than recurrent models. Towards this end, a new video prediction framework based on the MIMO manner is proposed in the field. We illustrate the error growth curve is remarkably reduced compared with existing work. 

Our contribution can be summarized in the following: (1) We reveal that the Multi-In-Multi-Out strategy plays a crucial role in addressing the long-standing error accumulation problem in video prediction for the first time. (2) We advocate a new MIMO architecture which leverages the Transformer-based architecture by exploiting the spatial context and local Spatio-temporal correlations of video data. A new decoder is designed to realize our MIMO idea. (3) Comprehensive experiments on four benchmarks are carried out to validate the effectiveness and fruitful insights are provided. Experimental results show that the proposed MIMO model achieves superior performance over the state-of-the-art SISO models in terms of accuracy and fidelity.

%% file: PaperBody/relatedwork4.tex
\subsection{Single-in-Single-Out Models.}

Mainstream video prediction models \cite{xingjian2015convolutional,shi2017deep,wang2017predrnn,wang2018predrnn++,oliu2018folded,wang2018eidetic,wang2019memory,guen2020disentangling,Yu2020EfficientAI,lin2020self,su2020convolutional,wu2021motionrnn,lee2021video} follow the typical recurrent strategy to synthesis the future frames, which takes single frame in the sequence as input and output the next one by the recurrent unit. As a prior work, Shi \emph{et al.}~\cite{xingjian2015convolutional} proposed ConvLSTM to address the problem. After that, its variants PredRNN \cite{wang2017predrnn} and MIM \cite{wang2019memory} are proposed which capture both spatial and temporal representations. MotionRNN \cite{wu2021motionrnn} and PhyDNet \cite{guen2020disentangling} explored different 
frameworks like GRU or two-branch methods to achieve promising results. However, it is still an inherent problem for them to capture long-term dependency and extrapolate longer sequences. 
To alleviate the pixel-level error propagation, a lot of work 
tried to learn the high-level semantic structures first, such as human pose~\cite{villegas2017learning}, segmentation maps~\cite{lee2021revisiting}, then translate these to pixels space. LMC-Memory\cite{lee2021video} proposed to predict long-term future frames by storing and recalling long-term motion contexts of video sequences. However, all of their error still accumulates at a compound rate. In contrast, our MIMO model fundamentally changes the error propagation way rather than alleviating the error per frame.\\
\subsection{Multi-in-Multi-Out Models.}
Instead of recurrent ways, MIMO models generate all future frames at once. It has been proved that these models \cite{bontempi2008long,taieb2010multiple} for multi-step-ahead time series prediction show superiority over the others. Compared with SISO, MIMO is able to capture the relationships of frames not only happened before but also happened afterwards. Few attempts proposed MIMO-based models to anticipate videos. For example, Liu \emph{et al.}~\cite{liu2017video} proposed DVF, which used a 3D convolution autoencoder to learn voxel flows and synthesize video frames by flowing pixel values from existing ones. FutureGAN~\cite{aigner2018futuregan} is also an encoder-decoder GAN model by stacking Spatio-temporal 3d convolutions, it could predict multiple future frames at once. SimVP~\cite{gao2022simvp} is a simple model that is completely built upon CNN. The MIMO predictive strategy of SimVP makes it achieve significant performance only with basic CNNs. However, they did not exploit the real strength of MIMO architectures. Thus we aim to develop a top-performing MIMO model to rise the attention for them in video prediction. \\
\subsection{Transformer Models for Videos.}
The advantage of modeling long-term dependency of videos enables Transformers~\cite{vaswani2017attention,dosovitskiy2020image,liu2021swin} to be widely applied in the video domain \cite{zhou2018end,gabeur2020multi,girdhar2019video,liu2020convtransformer,arnab2021vivit,dzabraev2021mdmmt,liu2021video,wang2021end,zhang2021token}.
For example, Zhou \emph{et al.} \cite{zhou2018end} proposed an end-to-end video captioning model based on Transformer.
Kim \emph{et al.} \cite{kim2018spatio} proposed a novel Spatio-temporal Transformer Network (STTN) for Video Restoration. The works in~\cite{gabeur2020multi,dzabraev2021mdmmt} investigated the potential of Transformers on video retrieval tasks.
Besides, recent advances~\cite{girdhar2019video,zhang2021token,arnab2021vivit,liu2021video} focused on designing the T
Transformer architectures for video classification.
VisTR~\cite{wang2021end} is a video instance segmentation framework which outputs the masks for each instance in parallel. However, the Transformer aiming at video prediction is still underdeveloped and our work fills this gap.


%% file: PaperBody/method2.tex
\subsection{Problem Formulation}
Suppose we observe a video sequence $\mathcal{S}_{t-m+1:t} := \{\mathcal{X}_{t-m+1},$ $\mathcal{X}_{t-m+2}, \dots, \mathcal{X}_t \}$, 
where $\mathcal{X}_t\in \mathbb{R}^{C_0\times H_0\times W_0}$ represents the frame (channel $C_0$, height $H_0$, width $W_0$) at the current time step $t$.
The goal of the Spatio-temporal sequence prediction is to generate the most likely length-$n$ sequence in the future given the previous length-$m$ sequence:
\begin{equation}
\begin{split}
\mathcal{S}^*_{t+1:t+n} = \mathop{\arg\max}\limits_{\mathcal{S}_{t+1:t+n}} \  p(\mathcal{S}_{t+1:t+m} | \mathcal{S}_{t-m+1:t})
\end{split}
\label{eq: Spatiotemporal Sequence Prediction}
\end{equation}
\noindent \textbf{SISO predictive strategy.} 
To solve the problem, the Single-In-Single-Out architectures are widely used. They can naturally preserve the time order logic of the generated frames. Their prediction paradigm can be written as:
\begin{equation}
\begin{split}
\mathcal{X}_{t'+1} = F(\mathcal{X}_{t'}, M_{t'})+\sigma_{t'},
\end{split}
\label{eq: RNN_solve}
\end{equation}
where $F$ denotes the model, $M_{t'}$ is the memory cell that remembers the past historical information. $\sigma_{t'}$ is a Tensor noise term. $\mathcal{X}_{t'}$ is the observed frame ($t'\leq t$) or the generated frame ($t'>t$).
Note that SISO predictive strategy is susceptible to the error accumulation problem.
For simplicity, consider an Auto Regressive with $K$=$1$ and without exogenous inputs model:
\begin{equation}
\begin{split}
\mathcal{X}_{t} = A\mathcal{X}_{t-1}+\sigma_{t}
\end{split}
\label{eq: AutoRegressive}
\end{equation}
where $A$ is a learnable Tensor. 
It is easy to have:
\begin{equation}
\begin{split}
\mathcal{X}_{t+n} = A^{t+n-1}\mathcal{X}_{t}+\sum_{k=t}^{t+n-1}A^k\sigma_{t+n-k}+\sigma_{t+n}
\end{split}
\label{eq: error}
\end{equation}
It is clear that the error $R_{t+n}=\sum_{k=t}^{t+n-1}A^k\sigma_{t+n-k}+\sigma_{t+n}$ will accumulate as time step $t+n$ increases, which makes accurate prediction challenging. Besides, the long-term dependency problem inherited from RNNs would further limit the performance of the SISO strategy on video prediction.

\begin{figure}[t]
	\centering
	\includegraphics[width=0.9\linewidth]{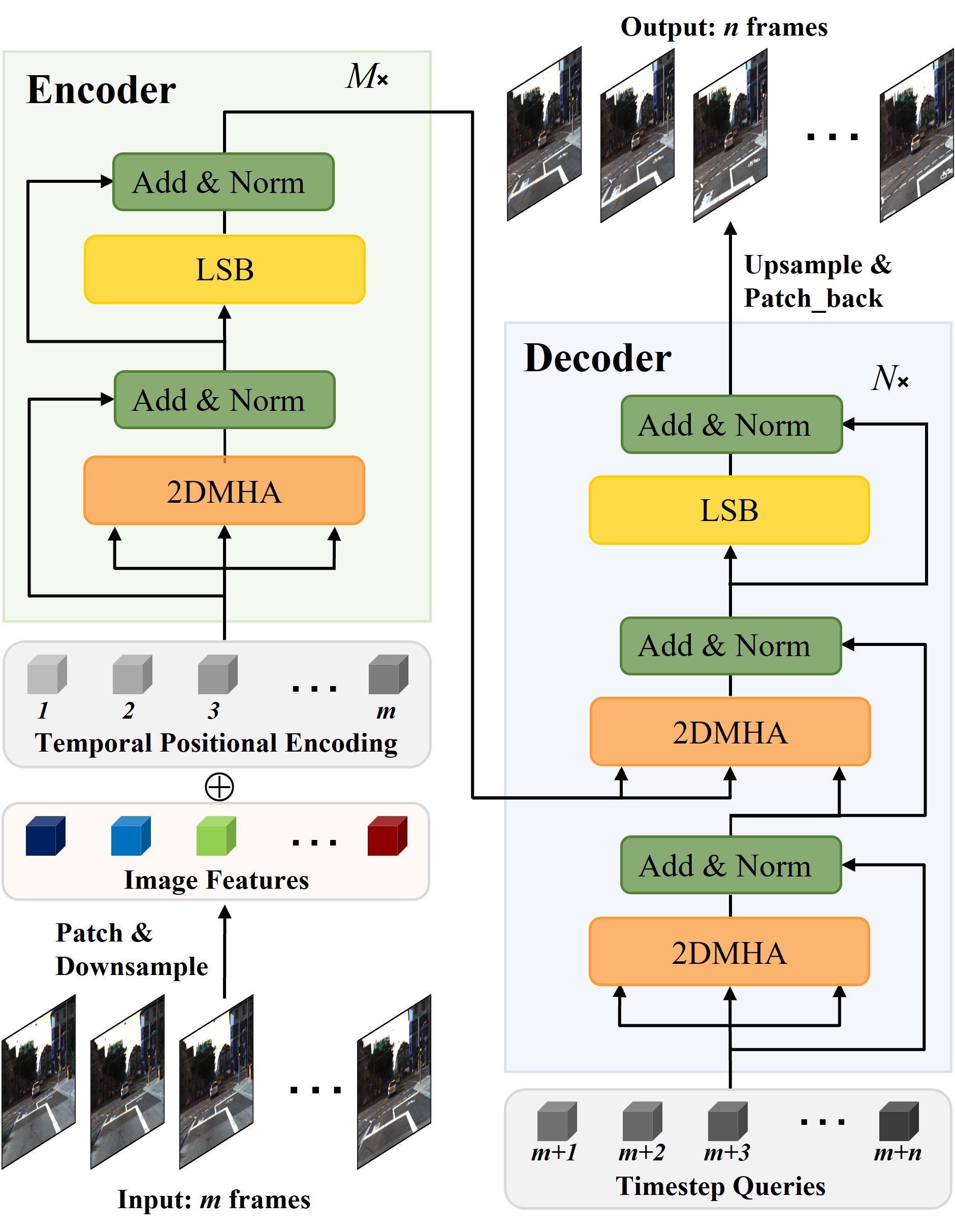}
	\caption{Illustration of MIMO-VP architecture. 
	}
	\label{fig:model}
\end{figure}

\noindent \textbf{MIMO predictive strategy.} 
There is an alternative way to deploy MIMO predictive strategy:
\begin{equation}
\begin{split}
\mathcal{S}_{t+1:t+n} = F(\mathcal{S}_{t-m+1:t})+ \Sigma_{t+1:t+n},
\end{split}
\label{eq: our_solve}
\end{equation}
where $\Sigma_{t+1:t+n}:=\{\sigma_{t+1},\dots,\sigma_{t+n}\}$ is the normal prediction error. Therefore, the MIMO strategy breaks the error accumulation by synthesising all the frames in parallel. However, it is not sufficient to deduce MIMO models always outperform SISO models. Actually, the single error $\sigma_{t+i}$ of one frame can be extremely high since the MIMO models may miss some important relationships which SISO models naturally capture. Notably, it is simple to present MIMO architectures but it is non-trivial to equip them with a strong ability to model long-term and short-term Spatio-temporal dependencies which are important for complex video data.

\subsection{Our MIMO-VP Architecture}
Encouraging by the great success of Transformer \cite{vaswani2017attention} models, we build up our MIMO architecture specific for video prediction based on the Transformer backbone. Their placeholder embedding method also helps to capture the global dependency of the frames. However, a pure Transformer is a Multi-In-Single-Out architecture and achieves only moderate performance in video prediction since they lack of ability to model complex Spatio-temporal dynamics. There is still a big gap in our goal to tailor it for video data. To tackle the issue, we propose a series of surgeries to enhance the ability of the model to capture spatial structures and local Spatio-temporal dependencies. Our final model is presented in \Cref{fig:model}.

The formulation of our full model can be written as:
\begin{equation}
\mathcal{S}_{t+1:t+n} = \Phi_d(\Phi_e(\mathcal{S}_{t-m+1:t}), \mathcal{T}_{t+1:t+n})
\label{eq: learning procedure}
\end{equation}
where $\Phi_e$ and $\Phi_d$ are the encoder and decoder modules respectively. $\mathcal{T}_{t+1:t+n}$ is a length-$n$ timestep queries. For an input video $\mathcal{S}_{t-m+1:t}\in \mathbb{R}^{m\times C_0\times H_0 \times W_0}$ with $m$ frames, we first decompose each frame into non-overlapping patches and then obtain the sequence-level feature maps $h\in \mathbb{R}^{m\times C \times H\times W}$ after several convolution layers.

Note that, we operate 2D Convolution Multi-Head Attention (MHA) first to generate the query map and paired key-value map of each frame. It benefits the model to simultaneously learn the temporal correlation and preserve the spatial information of sequence. Moreover, different from the existing auto-regressive prediction paradigm of Transformers,
we do not apply the mask operation on the multi-head self-attention in the decoder block. Because our decoder inputs are time step embedding that is already known. In our model, each generated frame is conditioned not only on the past sequence but also on its future frames, which preserves the dependency of future frames. After exploration, we figure out two operations can boost the performance drastically. The details are provided in the following. 

\noindent \textbf{Local Spatio-Temporal Block}
We figure out one reason that the vanilla Transformer is limited in performance since they lack the ability to capture the short-term variation of sequences which is crucial for video generation. Therefore, we are motivated to install a local Spatio-temporal block which aims to learn the high order relationship between time step queries $\mathcal{T}$ and their corresponding output sequence $\mathcal{S}$. We find out a simple way to implement this idea by installing this block in the place of the FCN layers of the encoder and decoder. Theoretically, this block can be any neural network which has the ability to capture Spatio-temporal information for video data. While a more complex design potentially will result in more gains, we hope to design this block as simple as possible to validate the importance of local Spatio-temporal clues for video prediction. Inspired by most MIMO frameworks in video domains, we choose a simple 3D Convolution layer which is widely used to capture these clues as our block. Specifically, the block is consisted of two embedding layers with each is a \texttt{3DConv}, \texttt{LayerNorm}, \texttt{SiLU} block. The kernel size of \texttt{3DConv} is $3\times 3\times 3$.

\noindent \textbf{Multi-Out Decoder}
To achieve our goal, we need to rectify the original decoder to enable it to produce several future frames parallelly. The challenge here is how to maintain the order information of in video sequences. Considering the permutation-invariant character of the Transformer architecture, we here focus on the temporal positional encoding of the sequence. Specifically, we define a time step vector $T=[1,2,\dots,m+n]$. Then we embed $T\in \mathbb{R}^{1\times (m+n)}$ to $\hat{T}\in \mathbb{R}^{(m+n)\times C}$ by an embedding layer. Finally, we expanse $\hat{T}$ to $\mathcal{T}\in \mathbb{R}^{(m+n)\times C\times H\times W}$ along the height and width dimensions. Accordingly, the temporal positional encoding of input sequence is $\mathcal{T}_{1:m} \in \mathbb{R}^{m\times C\times H\times W}$. After that, we regard $\mathcal{T}_{m+1:m+n}\in \mathbb{R}^{n\times C\times H\times W}$ as time step queries and input them to the decoder module. In this way, our new decoder is capable of dealing with multi-output scenarios. SISO strategy is somehow a specific case of MIMO strategy when the input length is one. The details of this Multi-Out decoder are illustrated in \textit{supplementary materials}.  

\begin{table*}[t]
	\small
	\centering
	\tabcolsep10pt
	
	\begin{tabular}{l|ccc|ccc}
		\hline
		\multirow{2}{*}{Method} & \multicolumn{3}{c|}{10 - 10}& \multicolumn{3}{c}{10 - 30} \\
		\cline{2-7}
		& MSE  $\downarrow$ & MAE  $\downarrow$ & SSIM $\uparrow$& MSE  $\downarrow$ & MAE  $\downarrow$ & SSIM $\uparrow$\\
		\hline
		\hline
		ConvLSTM \cite{shi2017deep} & 103.3& 182.9 & 0.707 &-&-&-\\
		MIM \cite{wang2019memory} & 44.2 & 101.1 & 0.910 &-&-&- \\
		CrevNet\cite{yu2019crevnet} & 22.3 & - & 0.949 &-&-&- \\ \hdashline
		ConvTTLSTM \cite{su2020convolutional}& 56.0& 92.9& 0.913 & 105.7  & - & 0.840\\
		PhyDNet \cite{guen2020disentangling} & 24.2  & 70.3 & 0.947&58.7 & 144.1 & 0.852 \\
		LMC-Memory \cite{lee2021video} & 41.5  & - & 0.924 &73.2 & - & 0.879\\
		MotionRNN \cite{wu2021motionrnn}& 29.2  & 87.7& 0.933&64.2 & 156.1 & 0.844\\ 
		SimVP \cite{gao2022simvp}& 23.8 & 68.9 & 0.948 &-&-&- \\
		\hline
		\textbf{MIMO-VP} & \textbf{17.7} & \textbf{51.6} & \textbf{0.964} & \textbf{31.6}& \textbf{79.1}& \textbf{0.934}\\
		\hline
	\end{tabular}
        \caption{Quantitative comparison of different models on Moving MNIST dataset.}
	\label{Tab: Moving MNIST}
\end{table*}

%% file: PaperBody/experiment3.tex
\subsection{Dataset Descriptions}
\begin{figure}[t]
	\centering
	\includegraphics[width=0.9\linewidth]{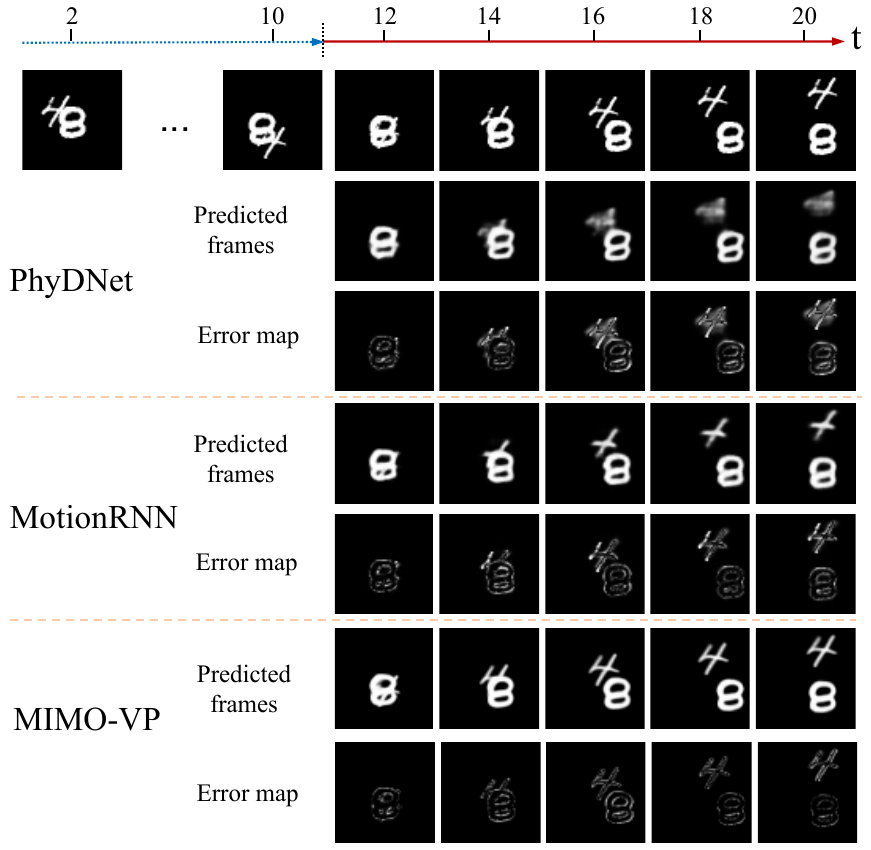}
	\caption{Prediction examples on Moving MNIST dataset. 
	}
	\label{fig:mnist}
\end{figure}

\begin{figure}[t]
	\centering
	\includegraphics[width=0.9\linewidth]{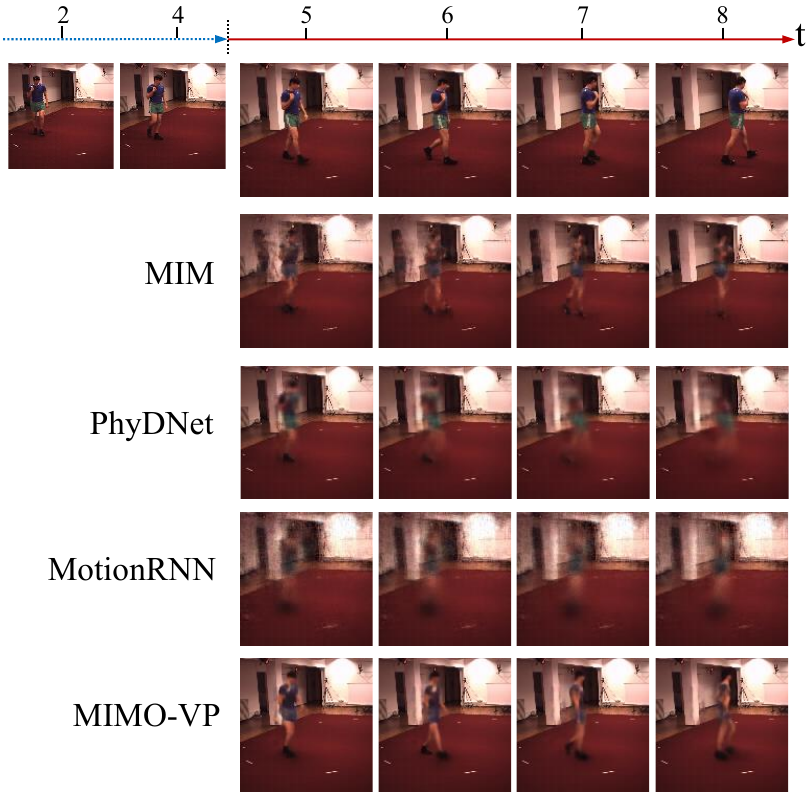}
	\caption{Prediction examples on Human3.6M dataset. 
	}
	\label{fig:human}
\end{figure}
\noindent \textbf{Moving MNIST} \cite{srivastava2015unsupervised} is a standard synthetic dataset for video prediction. We generate training set following \cite{guen2020disentangling},
and adopt the widely used testing set of 10,000 sequences provided by \cite{srivastava2015unsupervised}. 

\noindent \textbf{Human3.6M} \cite{ionescu2013human3} dataset is a real-world human pose dataset comprising 17 kinds of human actions and 3.6 million poses. We use subjects S1, S5, S6, S7 and S8 for training, and subjects S9, S11 for testing. We predict $4$ frames in the future conditioned on the $4$ observed frames.

\noindent \textbf{Weather} dataset is a real radar echo dataset for precipitation nowcasting, which is obtained from the local weather bureau. Our training set and testing set contain $9,600$ and $2400$ radar sequences (resolution 128$\times$128$\times$1) respectively.

\noindent \textbf{KITTI} dataset is one of the most popular datasets for autonomous driving and also a benchmark for computer vision algorithms. We predict $5$ frames in the future when taking $5$ observed frames as input.

\noindent \textbf{Implementation Details}
We use Adam \cite{kingma2014adam} optimizer with 0.0005 learning rate and L1$+$L2 loss to train our model.  We implemented the model with PyTorch and conducted experiments on NVIDIA V100 GPUs. Mean Squared Error (MSE), Mean Absolute Error (MAE), the Structural Similarity(SSIM), LPIPS~\cite{zhang2018unreasonable} and Peak Signal to Noise Ratio (PSNR) are evaluated in our work. All these metrics are averaged for all predicted frames. Additionally, we also take Critical Success Index (CSI) into account for Weather dataset. Details are contained in \textit{supplementary materials}.

\subsection{Quantitative and Qualitative Comparison}
\label{sec:IS MIMO beats SISO ?}
\noindent \textbf{Results on Moving MNIST.}
Firstly, we report the results which use 10 frames to predict 10 future frames. As shown in \Cref{Tab: Moving MNIST} and \Cref{fig:mnist}, our model significantly outperforms all the other models. Current RNN-based models get blurry and wrong predictions in this case, and their prediction error grows larger over time. On the contrary, our model can predict high quality frames. 
To further explore our performance on long-term prediction, we conduct an experiment by predicting 30 future frames conditioned on 10 input frames. As shown in \Cref{Tab: Moving MNIST}, our method far surpasses the others in predicting 30 frames in terms of the MSE metric. Compared with PhyDNet, our MIMO-VP improves by $46.2\%$ in MSE, which is much larger than $26.9\%$ in 10-10 prediction task. 

\begin{table}[t]
	\small
	\centering
	\tabcolsep 9pt
	{
	\begin{tabular}{l|ccc}
		\hline
		Method   & MAE/100 $\downarrow$ & SSIM $\uparrow$ & PSNR $\uparrow$ \\
		\hline 
		\hline
		MIM & 17.8& 0.790 & 20.57  \\
		PhyDNet & 16.2 & 0.901 &-\\
		LMC-Memory  & 17.0 & 0.820 &21.05\\
		MotionRNN  & 14.8& 0.846 & 22.16   \\
		\textbf{MIMO-VP}  & \textbf{10.4} & \textbf{0.941}  & \textbf{24.52} \\
		\hline
	\end{tabular}}
	\caption{Quantitative comparisons on Human3.6M
	dataset.}
	\label{Tab:human}
\end{table}

\begin{table}[t]
\small
\centering
\tabcolsep3pt
  \begin{tabular}{l|cccc}
  \hline
    Method   & MAE/100 $\downarrow$ & SSIM $\uparrow$& LPIPS $\downarrow$ & PSNR $\uparrow$ \\
	\hline 
	\hline
	CrevNet & 56.3 & 0.587  & 0.454  & 17.046  \\
		PhyDNet   & 44.9 & 0.674  & 0.403& 19.159 \\
		LMC-Memory  & 44.6 & 0.660  &  0.410  & 18.692\\
		MotionRNN  & 45.0 & 0.652  & 0.384& 18.931 \\
		\textbf{MIMO-VP}  & \textbf{40.9} & \textbf{0.703} & \textbf{0.308}& \textbf{19.616}  \\
     \hline
	\end{tabular}
\caption{Quantitative comparisons on KITTI dataset.}
\label{Tab: kitti}
\end{table}

\begin{table}[!]
\centering
\tabcolsep8pt
  \small
 {
  \begin{tabular}{l|cccc}
  \hline
    Method  & MAE $\downarrow$ & SSIM $\uparrow$ & CSI-20 $\uparrow$ & CSI-30 $\uparrow$\\
	\hline 
	\hline
	MIM  & 402.6 & 0.725& 0.402  & 0.164\\
		PhyDNet & 280.1 & 0.859 & 0.557 & 0.368 \\
		MotionRNN & 336.9 & 0.809 & 0.519 & 0.290 \\
		\textbf{MIMO-VP}  & \textbf{235.6} & \textbf{0.865} & \textbf{0.595} & \textbf{0.400}\\ 
     \hline
	\end{tabular}}
\caption{Quantitative comparisons on Weather dataset.}
\label{Tab: weather}
\end{table}

\begin{figure}[t]
	\centering
	\includegraphics[width=0.95\linewidth]{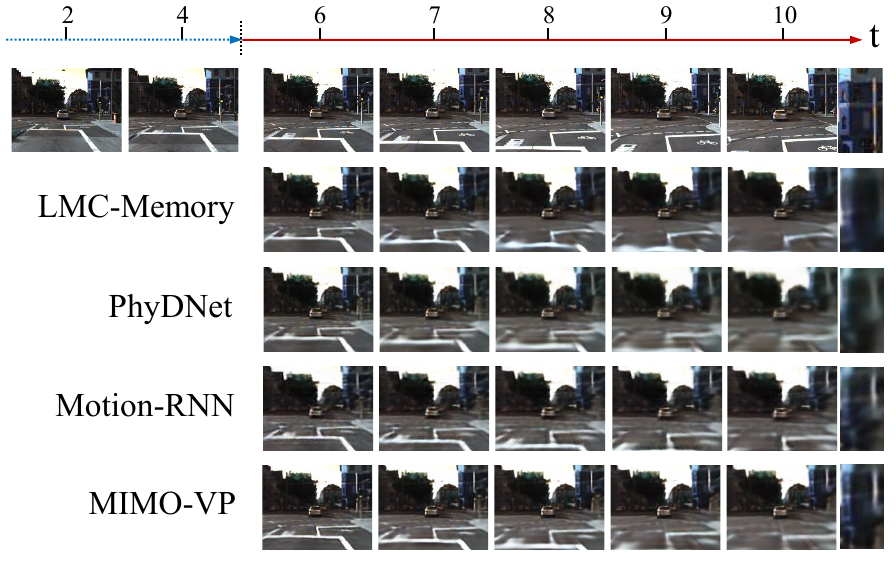}
	\caption{Prediction examples on KITTI dataset. 
	}
	\label{fig:KITTI}
\end{figure}

\begin{figure}[t]
	\centering
	\includegraphics[width=0.95\linewidth]{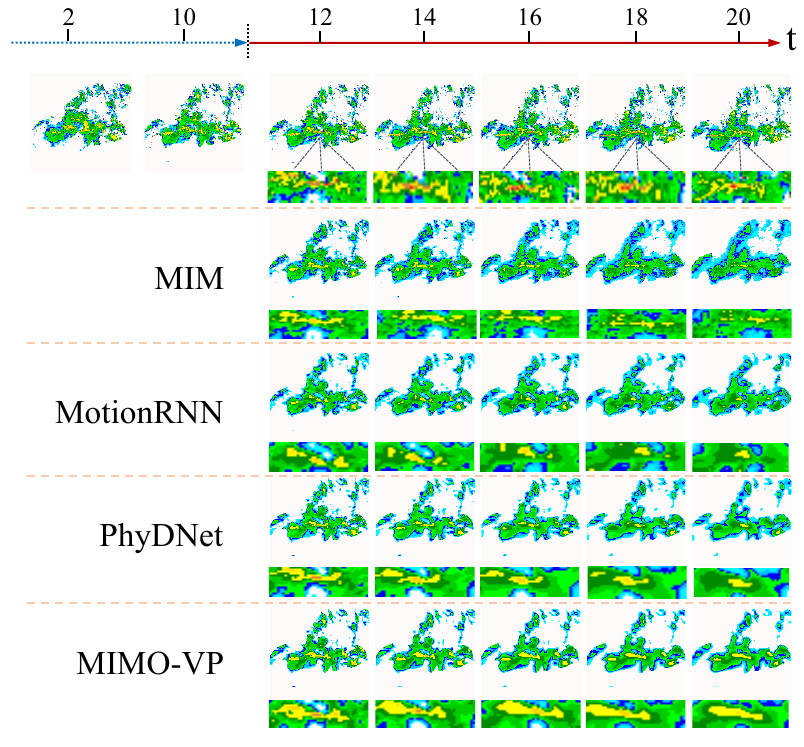}
	\caption{Prediction examples on Weather dataset. 
	}
	\label{fig:weather}
\end{figure}

\begin{table}[t]
\small
\centering
\tabcolsep3pt
  \begin{tabular}{l|cccc}
  \hline
    Method   & MSE $\downarrow$ & MAE $\downarrow$ & SSIM $\uparrow$ & \#Params (M)\\
	\hline 
	\hline
    CrevNet & 22.3 & - & 0.949  &- \\
	\hline	
	3D convolution  & 32.8 & 84.4 & -  & 19.8 \\
		ViT & 50.2 & 126.5 & 0.887  & 17.5\\
		Swin-Transformer & 28.8 & 77.2 & 0.940 & 21.8 \\
		SimVP & 23.8& 68.9 & 0.948 & 15.8\\
		\textbf{MIMO-VP} & \textbf{17.7}  & \textbf{51.6} & \textbf{0.964}  & 30.2 \\
     \hline
	\end{tabular}
\caption{Results of MIMO models on Moving MNIST.}
\label{Tab: MIMO_compare}
\end{table}

\noindent \textbf{Results on Human3.6M.}
\Cref{Tab:human} shows that MIMO-VP achieves the best performance on Human3.6M. 
Compared with PhyDNet, MIMO-VP improves by $35.8\%$ in MAE. 
In \Cref{fig:human}, complicated background and human motion give much more challenge for predicting accurate and reasonable results. While our MIMO-VP can still product relatively precise future frames comparing with other models.


\noindent \textbf{Results on KITTI.}
As shown in \Cref{Tab: kitti}, our proposed MIMO-VP achieves consistent improvement in all evaluation metrics. 
Compared with LMC-Memory, our model improves by $8.3\%$ in MAE.
Quantitative visualization results in \Cref{fig:KITTI} show that our model can preserve much more details of both background and foreground. 
For instance, from the right part of last predicted frames, we can find that the results of our MIMO-VP is much similar to GroundTruth while other methods produce obviously blurry frames. 

\noindent \textbf{Results on Weather.}
From Table \ref{Tab: weather}, we can find that our MIMO-VP has significantly improvement in CSI scores within different thresholds, which indicates our approach can predict more credible radar echoes for precipitation nowcasting. 
The visualization radar echo maps are shown in \Cref{fig:weather}. Different color represents different radar echo intensity (dBZs), \ie \ yellow indicates higher intensity than green.
Additionally, the center region of the radar echo is zoomed in and placed below each predicted frame. 
We find that our MIMO-VP predicts more correct yellow regions  than other methods, which means MIMO-VP is a more suitable method for precipitation nowcasting.

\noindent \textbf{The comparison of different MIMOs}
Quantitative comparisons are detailed in Table~\ref{Tab: MIMO_compare}, where 3D convolution is a fully convolutional network. 
For ViT \cite{dosovitskiy2020image} and Swin-Transformer \cite{liu2021swin}, we replace the classification layer with convolution operation to be consistent with original image size.  
We observe that all other MIMO methods can not achieve state-of-the-art performance when compared with the best SISO method--CrevNet \cite{yu2019crevnet} except ours. 
To verify the effectiveness of our MIMO-VP, we carry out an ablation study to analyse each key component.

\begin{table}[t]
	\small
	\centering
	\tabcolsep15pt
	\begin{tabular}{l|cc}
		\hline
		Method  & MSE $\downarrow$ & MAE $\downarrow$  \\
		\hline  
		\hline
		Without 2DMHA & 32.8 & 84.4  \\
		Without LSB  & 21.2 & 59.2  \\
		MISO-VP  &22.8 & 62.8 \\
		\hline
		\textbf{MIMO-VP}  & \textbf{17.7} & \textbf{51.6}  \\
		\hline
	\end{tabular}
	\caption{Ablation Study for Each Components.}
	\label{Tab: ablation}
\end{table}

\begin{figure}[t]
	\centering
	\begin{subfigure}{0.491\linewidth}
		\centering
		\includegraphics[width=1\linewidth]{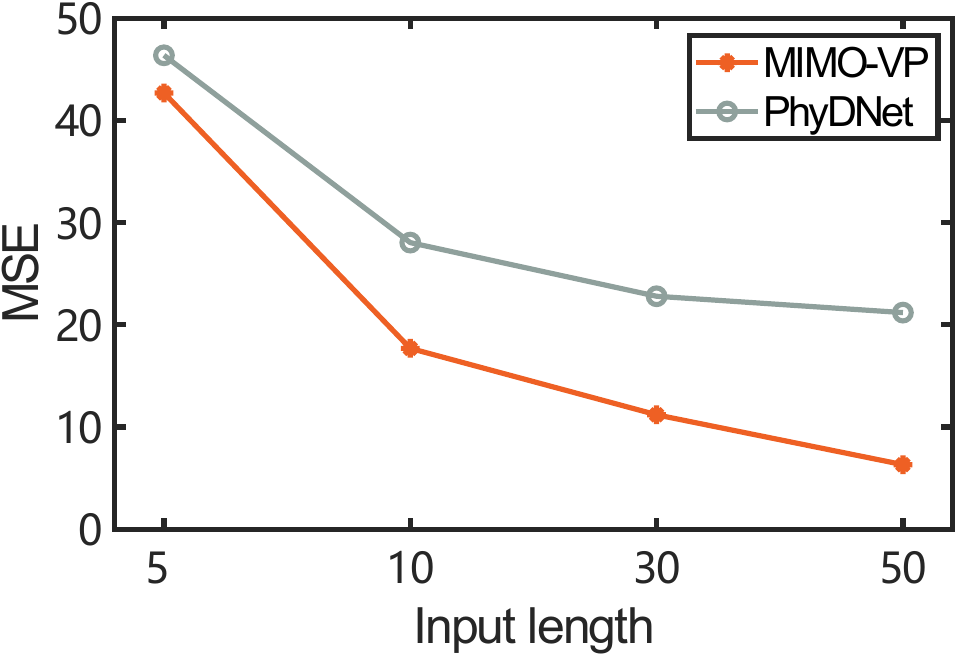}
		\caption{MSE w.r.t input length}
		\label{figure:long-term}
	\end{subfigure}
	\hfill
	\begin{subfigure}{0.491\linewidth}
		\centering
		\includegraphics[width=1\linewidth]{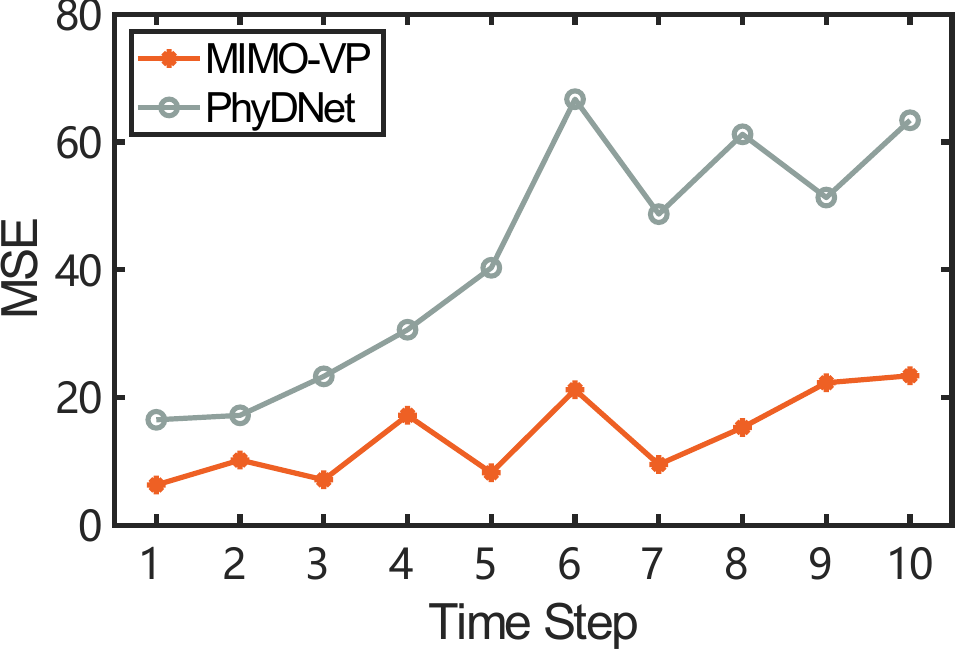}
		\caption{Frame-wise MSE}
		\label{figure:error accumulation}
	\end{subfigure}
	\caption{(a) Comparison of MIMO-VP and PhyDNet by predicting
10 future frames conditioned on different length of input sequence;
(b) Frame-wise MSE of MIMO-VP and PhyDNet on Moving
MNIST dataset.}
\end{figure}

\begin{figure}[t]
	\centering
	\begin{subfigure}{0.491\linewidth}
		\centering
		\includegraphics[width=1\linewidth]{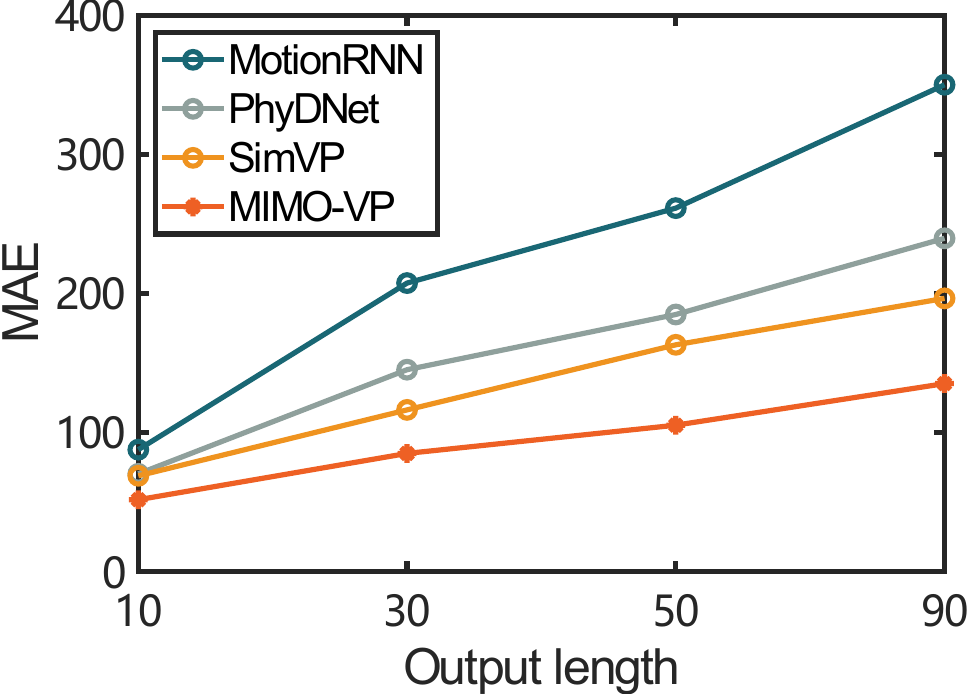}
		\caption{MAE w.r.t Output length}
		\label{figure:long term prediction1}
	\end{subfigure}
	\hfill
	\begin{subfigure}{0.491\linewidth}
		\centering
		\includegraphics[width=1\linewidth]{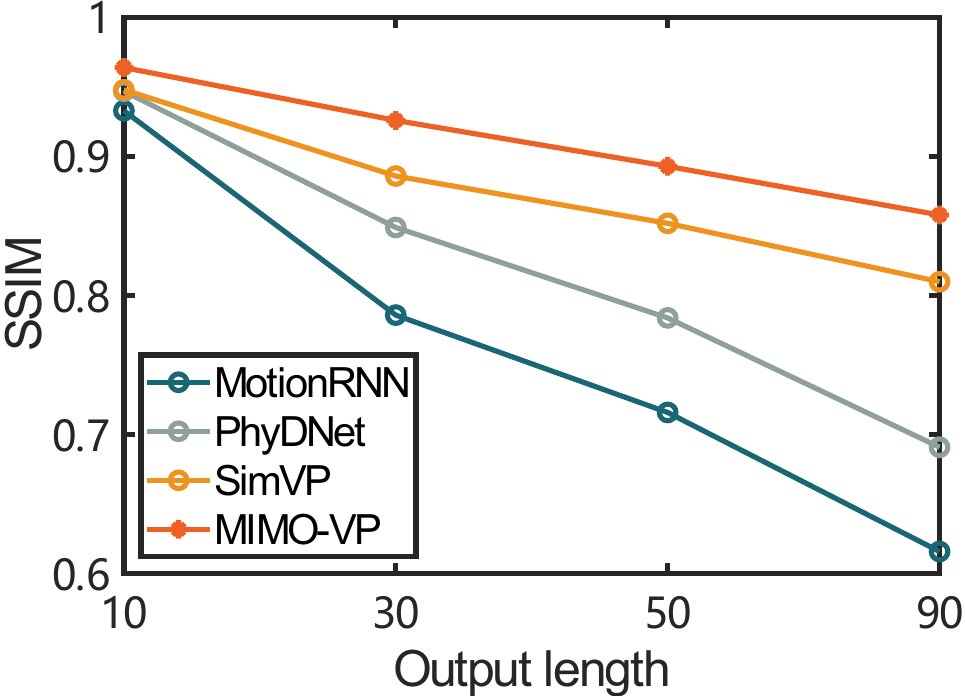}
		\caption{SSIM w.r.t Output length}
		\label{figure:long term prediction2}
	\end{subfigure}
	\caption{Results of long-term predication on Moving MNIST dataset.}
\end{figure}

\begin{table}[!t]
    \small
	\centering
	\tabcolsep10pt
	\begin{tabular}{l|ccc}
		\hline
		Method  & MSE $\downarrow$ & MAE $\downarrow$ & SSIM $\uparrow$ \\
		\hline 
		\hline
		MIMO-VP\_F & 21.6 & 59.4 & 0.955 \\
		\textbf{MIMO-VP} & \textbf{17.7} & \textbf{51.6} & \textbf{0.964} \\
		\hline
	\end{tabular}
	\caption{Comparison of MIMO-VP and its variant on Moving MNIST dataset.}
	\label{Tab: dependncy of future}
\end{table}


\subsection{Ablation Study of Each Component}
\label{sec:Which MIMO?}
 Quantitative results on Moving MNIST  dataset are shown in \Cref{Tab: ablation}. 
(1) \textbf{The Spatial Dependency of Frames}. We can see that MIMO-VP without 2DMHA results in a large performance degradation in comparison with MIMO-VP, with an increase of 15.1 and 32.8 in terms of MSE and MAE respectively. This phenomenon indicates that
2DMHA mechanism, which is able to capture the global long-term spatiotemporal relationships of sequences, makes an important contribution to the accurate video prediction.
(2) \textbf{The Local Spatio-temporal Block (LSB)}. The results of MIMO-VP without LSB also show inferior to those of MIMO-VP, which verifies that the local spatial-temporal variations of sequence are helpful for prediction.
(3) \textbf{The Multi-Out Decoder}. 
MISO-VP denotes a variant of MIMO-VP that adopts the MISO predictive strategy similar to vanilla Transformer. 
It infers the next one frame using the predicted frames, in a recursive manner.
Experimental results show that MIMO-VP beats down the MISO-VP with performance improves by $18.0\%$ and $14.0\%$ in MSE and MAE respectively. The reason may be that recursive single-out predictive strategy will trigger the error accumulation problem and overlook the stochastic dependency among the future frames.

\subsection{Comparison of MIMO and SISO}
\label{sec:Why MIMO beats SISO?}
\noindent \textbf{Significance of `MI’: long-term dependency }
Comparison results between MIMO-VP and PhyDNet of predicting 10 future frames on Moving MNIST with different input lengths are shown in \Cref{figure:long-term}.
It is observed that the MSE metric of MIMO-VP drops quickly with the input length increasing from 5 to 50. PhyDNet improves
significantly when the input length increases from 5 to 10,
while improves slightly from 10 to 50. Precisely, the MSE
of MIMO-VP and PhyDNet deceases by $64.3\%$  and $24.4\%$
respectively from 10 input frames to 50 input frames. Those
results demonstrate that our MIMO-VP has a stronger ability
to capture the long-term sptiotemporal dependency of
video sequences for accurate prediction, while PhyDNet, an
RNN-based method, suffers from the long-term dependency
problem because of the limited capacity of memory.

\noindent \textbf{Significance of `MO': free of error accumulation}
\Cref{figure:error accumulation} illustrates the frame-wise MSE of MIMO-VP and PhyDNet, corresponding to the sample in \Cref{fig:mnist}. 
We can observe that the curve of PhyDNet increase dramatically over time because PhyDNet suffers from the error accumulation problem. 
By contrast, the curve of MIMO-VP increase slightly because the normal predictive error, which is related to the digits overlapping, increases over time. 
To futher verify that the error accumulation problem is even worse in long-term case,  we compare the results of predicting 10, 30, 50 and 90 frames on Moving MNIST dataset. 
In training phase, all the models are trained by using 10 frames to predict next 10 frames, while in testing phase, 10, 30, 50, and 90 frames are predicted in a recurrent manner. 
Experimental results are shown in \Cref{figure:long term prediction1} and \Cref{figure:long term prediction2}, from which we can observe that the error accumulation problems of SISO models (MotionRNN, PhyDNet) is much severe than MIMO models (SimVP, MIMO-VP).  
Thus, `MO' models can overcome the error accumulation problem in video prediction. 

\noindent \textbf{Future frame dependency }
We compare our MIMO-VP with its variant \textit{MIMO-F}, a baseline which removes the self-attention module in the decoder so that it can not capture all the dependency of frames. 
Predictive results are shown in
\Cref{Tab: dependncy of future}, from which we can see MIMO-VP achieves much better performance than \textit{MIMO-F}.  This phenomenon verifies the benefit of MIMO-VP in preserving the dependency among the future frames for accurate prediction.


%% file: PaperBody/conclusion3.tex
In this paper, we propose a new MIMO architecture for video prediction, which overcomes the long-term dependencies and error accumulation problems inherently introduced in the Single-In-Single-Out (SISO) models. The results show that our model wins 1st place in many evaluations and challenge datasets and surpasses the state-of-the-arts SISO models in a large gap, especially over a longer period. By this work, we encourage the researchers in prediction area to shift more attention on MIMO models. We believe our model could serve as a new backbone and facilitate the future research of video prediction.
